\documentclass[conference]{IEEEtran}
\IEEEoverridecommandlockouts

\usepackage{cite}
\usepackage{amsmath,amssymb,amsfonts}
\usepackage{graphicx}
\usepackage{booktabs}
\usepackage{multirow}
\usepackage{array}
\usepackage{tabularx}
\usepackage{url}
\usepackage{xcolor}
\usepackage{stfloats}
\usepackage{tikz}
\usepackage{newtxtext}
\usepackage{newtxmath}
\usepackage{placeins}
\usetikzlibrary{arrows.meta,positioning,fit,calc}
\graphicspath{{figures/}}

\begin{document}

\title{Leakage-Controlled Structure-Preserving Augmentation and Template-Aware Evaluation for Bangla Symptom-Based Disease Prediction}

\author{
\IEEEauthorblockN{Sabah Nushra\IEEEauthorrefmark{1},
Shovan Samanta Turzo\IEEEauthorrefmark{2},
Moinul Hossain\IEEEauthorrefmark{3},
Md Kishor Morol\IEEEauthorrefmark{4}}
\IEEEauthorblockA{\IEEEauthorrefmark{1}Islamic University of Technology, Gazipur, Bangladesh --- sabahnushra@iut-dhaka.edu}
\IEEEauthorblockA{\IEEEauthorrefmark{2}Daffodil International University, Dhaka, Bangladesh --- turzo15-5785@diu.edu.bd}
\IEEEauthorblockA{\IEEEauthorrefmark{3}American International University--Bangladesh, Dhaka, Bangladesh --- 20-43120-1@student.aiub.edu}
\IEEEauthorblockA{\IEEEauthorrefmark{4}ELITE Research Lab, New York, USA --- kishormorol@ieee.org}
}

\maketitle

\begin{abstract}
Bangla symptom-based disease prediction is commonly evaluated on a small, imbalanced structured benchmark containing many profiles that differ by only one or two symptoms. We test whether train-only asymmetric Bernoulli perturbation improves classification using 725 unique profiles spanning 85 diseases and 172 binary symptoms. Experiments use ten repeated real-profile holdouts and parent-grouped inner cross-validation. The $2 \times 2$ design separates augmentation from hyperparameter tuning, with exact duplication, random oversampling, class weighting, and SMOTEN as balancing controls. For Bernoulli Naive Bayes (BNB), the gain over tuned raw training does not withstand Holm correction, although BNB attains the highest mean accuracy (97.50\%). The boosted-tree models benefit substantially: XGBoost, LightGBM, and CatBoost improve by 18.10, 28.45, and 10.09 percentage points, respectively. A Hamming-distance audit places 91.0\% of standard held-out profiles within one symptom bit of the training data. Accuracy decreases on a same-disease 1-bit-purged challenge; nevertheless, XGBoost and LightGBM retain large augmentation gains. We also evaluate robustness to missing and spurious symptoms and the faithfulness of BNB explanations. Overall, Bernoulli perturbation is useful for some model families, but benchmark accuracy should be interpreted together with template proximity.
\end{abstract}

\begin{IEEEkeywords}
Bangla disease prediction, data leakage, synthetic augmentation, class imbalance, template proximity, repeated evaluation, robustness, interpretability
\end{IEEEkeywords}

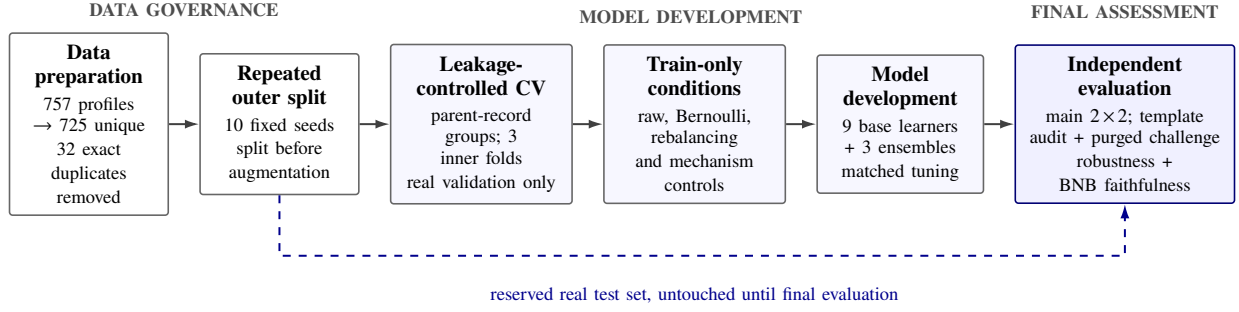
\begin{figure*}[!t]
\centering
\begin{tikzpicture}[
    font=\footnotesize,
    >=Latex,
    box/.style={draw=black!60, line width=.6pt, rounded corners=1.2pt,
                align=center, minimum height=13.5mm, inner xsep=1.5mm, inner ysep=1.6mm,
                fill=white},
    dev/.style={box, fill=blue!3},
    final/.style={box, fill=blue!6, draw=blue!45!black},
    flow/.style={-{Latex[length=1.9mm,width=1.2mm]}, line width=.75pt, draw=black!75},
    test/.style={-{Latex[length=1.9mm,width=1.2mm]}, dashed, line width=.75pt, draw=blue!55!black},
    small/.style={font=\scriptsize, text=black!72},
    label/.style={font=\scriptsize\bfseries, text=black!70}
]

\node[box, text width=18mm] (prep) {\textbf{Data preparation}\\[2pt]\scriptsize 757 profiles $\rightarrow$ 725 unique\\32 exact duplicates removed};
\node[box, text width=18mm, right=4mm of prep] (split) {\textbf{Repeated outer split}\\[2pt]\scriptsize 10 fixed seeds\\split before augmentation};
\node[dev, text width=21mm, right=4mm of split] (cv) {\textbf{Leakage-controlled CV}\\[2pt]\scriptsize parent-record groups; 3 inner folds\\real validation only};
\node[dev, text width=21mm, right=4mm of cv] (train) {\textbf{Train-only conditions}\\[2pt]\scriptsize raw, Bernoulli, rebalancing\\and mechanism controls};
\node[dev, text width=19mm, right=4mm of train] (models) {\textbf{Model development}\\[2pt]\scriptsize 9 base learners + 3 ensembles\\matched tuning};
\node[final, text width=26mm, right=4mm of models] (eval) {\textbf{Independent evaluation}\\[2pt]\scriptsize main $2\!\times\!2$; template audit + purged challenge\\robustness + BNB faithfulness};

\coordinate (flowlevel) at (prep.center);
\draw[flow] (prep.east |- flowlevel) -- (split.west |- flowlevel);
\draw[flow] (split.east |- flowlevel) -- (cv.west |- flowlevel);
\draw[flow] (cv.east |- flowlevel) -- (train.west |- flowlevel);
\draw[flow] (train.east |- flowlevel) -- (models.west |- flowlevel);
\draw[flow] (models.east |- flowlevel) -- (eval.west |- flowlevel);

\coordinate (testlevel) at ($(split.south)+(0,-8mm)$);
\draw[test] (split.south) -- (split.south |- testlevel)
                        -- (eval.south |- testlevel) -- (eval.south);
\node[small, text=blue!55!black, anchor=north] at ($(train.south)+(0,-9.8mm)$)
{reserved real test set, untouched until final evaluation};

\node[label, anchor=south] at ($(prep.north)!0.5!(split.north)+(0,2.2mm)$) {DATA GOVERNANCE};
\node[label, anchor=south] at ($(cv.north)!0.5!(models.north)+(0,2.2mm)$) {MODEL DEVELOPMENT};
\node[label, anchor=south] at ($(eval.north)+(0,2.2mm)$) {FINAL ASSESSMENT};

\end{tikzpicture}
\caption{Experimental pipeline. Real profiles are deduplicated and split before any augmentation. Parent-grouped cross-validation, augmentation, and model fitting use only the training partition. The reserved real test partition bypasses model development and is used only once, for final evaluation.}
\label{fig:pipeline}
\end{figure*}

\section{Introduction}
\label{sec:intro}

Symptom-based disease prediction is a practical low-resource classification problem because early decision-support research can be conducted with structured symptom profiles rather than large clinical text corpora. The issue is particularly pertinent for Bangla, for which publicly available clinical language resources are limited. In this study, 172 binary symptom indicators are encoded in a structured benchmark containing 85 disease labels~\cite{zannat_dataset}. Following exact deduplication, 725 distinct profiles remain, and several diseases have only a handful of examples. These records are disease--symptom profiles rather than longitudinal patient episodes, so high benchmark accuracy cannot be equated with evidence of clinical effectiveness.

Evaluation is fragile for small structured datasets. Synthetic augmentation can cause information leakage if variants are generated before the train--test split or if descendants of one real profile are placed in different validation folds~\cite{kaufman_leakage}. A second confound arises when hyperparameters are optimized differently across training conditions, making augmentation effects difficult to separate from model-selection effects. A third challenge is that many profiles contain a similar number of symptoms and differ by only a few bits, so closely related templates may be distributed across both the training and test sets. High test accuracy can therefore reflect local template similarity rather than broader out-of-sample generalization.

The issue of the large label space and the limited amount of real evidence per class exacerbates these issues. The apparent ranking of classifiers will vary depending on small changes in preprocessing, model selection or split composition because there are few real profiles per disease.

We study an asymmetric Bernoulli perturbation designed for binary symptom vectors. The generator retains observed symptoms with high probability and activates absent symptoms only rarely, so all generated samples remain in the binary feature space. We do not treat the resulting synthetic profiles as additional clinical evidence; instead, they provide local regularization of the training data. We measure their impact using a balanced $2\times 2$ experiment, mechanism controls, repeated statistical tests, a Hamming-distance audit, a more stringent template-purged challenge, and robustness tests with missing and spurious symptoms.

The analysis addresses four questions: Is augmentation helpful after matched tuning? Do gains on conventional holdouts arise from duplication or class balancing? How strongly does conventional holdout performance depend on template proximity? How do the trained models behave when symptoms are corrupted? Together, these analyses provide a conservative and reproducible assessment of structure-preserving augmentation for the Bangla disease--symptom benchmark.

\section{Literature Review}
\label{sec:literature}

\subsection{Bangla Symptom-Based Disease Prediction}
Previous studies of Bangla disease prediction have used both structured symptom matrices and free-text descriptions. Zannat \emph{et al.}~\cite{zannat_ecce} evaluated classical machine-learning models and voting ensembles on the public Bangla disease--symptom matrix~\cite{zannat_dataset}. Hossain \emph{et al.}~\cite{hossain} used transformer-based classification for Bengali symptom text, while Biswas and Das~\cite{biswas} used a CNN-based symptom classifier. These studies demonstrate the feasibility of Bangla disease prediction, but they differ in input representation and evaluation design. Because the public structured benchmark records only the presence or absence of each symptom, credible evaluation requires careful data splitting, model tuning, and control of near-duplicate profiles.

Previous studies also have shown that the importance of representation can be equal to that of the choice of classifiers. Language-model analysis was used by Hassan~\emph{et al.}~\cite{hassan} to achieve good symptom-based classification, and richer disease--symptom relations can be captured by a graded resource like DSR~\cite{dsr} than can be captured by a binary matrix. Additional structured clinical attributes are also shown to be advantageous by NLICE~\cite{nlice}. The results urge caution in using binary benchmarks: augmentation can regularize the observed feature space, but cannot recreate severity, duration, temporal order, or any other clinical variable not collected.

\subsection{Augmentation for Discrete Clinical Features}
Small clinical datasets may require rebalancing or regularization. SMOTE~\cite{smote} interpolates minority samples, which can produce fractional values for binary symptom indicators. Synthetic-record generation methods such as NLICE~\cite{nlice} and MedGAN~\cite{medgan} avoid this specific mismatch. The perturbation considered here has a more limited objective: it creates small binary variations of training profiles and is compared with exact duplication, random oversampling, class weighting, and fold-local SMOTEN.

\subsection{Leakage and Benchmark Reliability}
When held-out data influence preprocessing, resampling, or model selection, performance estimates can be inflated~\cite{kaufman_leakage}. Model-selection bias is also well known when tuning and evaluation are not clearly separated~\cite{varma_simon}. When augmentation is used, synthetic samples should therefore be generated only from the current training partition, and inner-validation folds should preserve parent groups. Exact deduplication is also insufficient when binary profiles differ by only one or two symptoms. For these reasons, we use split-first augmentation, matched tuning, repeated holdouts, train--test Hamming audits, and a same-disease purged challenge.

\section{Methodology}
\label{sec:method}

Figure~\ref{fig:pipeline} shows the full experimental workflow. The design follows one rule: real profiles are split before any synthetic data are generated, and all augmentation and model selection take place inside the outer-training set.

\subsection{Dataset Preparation and Repeated Holdout Design}
The source workbook contains 757 curated disease--symptom profiles. After removing 32 exact duplicate rows, there are 725 unique profiles (2 to 41 per class) that identify 85 diseases. A fixed preprocessing rule sets three missing symptom entries to 0. In addition, two identical 172-bit symptom vectors carry different disease labels, making those classes impossible to separate perfectly in this representation.

For each outer seed $s$, the real profiles are partitioned within every disease class. If class $c$ has $n_c$ profiles, its held-out count is
\begin{equation}
 n_{\mathrm{test},c}=\min\{n_c-1,\max(1,\operatorname{round}(0.15n_c))\}.
\end{equation}
This rule gives 609 real training profiles and 116 real test profiles in every outer split, with all 85 diseases present in both partitions. The ten fixed seeds are $\{11,22,33,42,55,66,77,88,99,111\}$.

For each outer training set, we create three folds using stable real-record IDs. Fold grouping is based on the \emph{original parent record ID}, not on the disease label. For a class with only one remaining parent, that parent is kept in the inner-training portion; parents from other classes can appear in validation folds. Only real records are used for validation. Every synthetic child inherits its parent ID, so a parent and its children never appear on opposite sides of an inner training--validation split. The outer test set is not used for model selection, augmentation selection, or ensemble weighting.

\subsection{Bernoulli Symptom Perturbation}
For a binary parent vector $x\in\{0,1\}^{172}$, every symptom dimension of a synthetic child $\tilde{x}$ is sampled independently as
\begin{equation}
P(\tilde{x}_j=1\mid x_j)=
\begin{cases}
p, & x_j=1,\\
q, & x_j=0,
\end{cases}
\qquad p=0.90,\;q=0.005.
\end{equation}
Each observed symptom is therefore retained with high probability, while each absent symptom is activated only rarely. The generator applies an asymmetric local perturbation rather than producing an unbiased copy of its parent.

Each disease contributes fifty synthetic children, so the outer-training set contains $609+85\times50=4{,}859$ rows. A child differs from its parent by a mean (median) of 1.35 (1.0) bits, corresponding to a normalized Hamming distance of 0.0079, whereas two real profiles of the same disease differ by 2.37 bits on average. The synthetic samples are deliberately local and are used only for regularization and class balance.

Sensitivity analysis uses only inner cross-validation over $p\in\{.80,.85,.90,.95\}$, $q\in\{0,.0025,.005,.01\}$, and $n\in\{10,25,50,100\}$ on three seeds. The setting ($.90,.005,50$) yields the highest mean macro-F1 for both BNB and logistic regression. As a discrete fold-local control, singleton classes in each inner-training fold are duplicated once before applying SMOTEN with $k=1$; validation data remain real. Numeric ADASYN is excluded because interpolation can leave the binary support $\{0,1\}^{172}$.

\subsection{Models and Controlled Comparisons}
The evaluated base learners include a cosine-similarity nearest-prototype classifier, logistic regression (LR), Bernoulli Naive Bayes (BNB), cosine k-nearest neighbors (KNN), random forest (RF), multilayer perceptron (MLP), XGBoost~\cite{xgboost}, LightGBM~\cite{lightgbm}, and CatBoost~\cite{catboost}. We also evaluate three ensemble variants: soft voting (Ens-Soft), hard voting (Ens-Hard), and voting weighted by inner-cross-validation scores (Ens-Weighted).

The main experiment has 4 conditions, raw/untuned, raw/tuned, Bernoulli/untuned and Bernoulli/tuned. The two pipelines (raw pipeline and Bernoulli pipeline) have the same search spaces, so that the effect of augmentation can be separated from the effect of choosing hyperparameters. Added as balancing controls are Exact duplication, Random oversampling (ROS), class weighting (for models supporting it) and fold-local SMOTEN. The mechanism analysis is done separately for each of these three perturbations: removal only $p=.90,q=0$; addition only $p=1,q=.005$; full Bernoulli $p=.90,q=.005$.

The primary metrics are accuracy and macro-F1; secondary metrics include macro precision, macro recall, weighted F1, MCC, Cohen's $\kappa$, and one-vs-rest macro ROC--AUC. Stability is reported as mean$\pm$SD across the ten outer splits. The primary paired comparison is between Bernoulli-tuned and raw-tuned training. Wilcoxon signed-rank tests assess split-level differences, with Holm correction applied across models. For each split, we also calculate Wilson intervals and exact McNemar tests.

\subsection{Template-Aware and Robustness Evaluation}
For each standard split, we compute the Hamming distance from every held-out profile to the real training set. We then construct a harder \emph{1-bit-purged same-disease challenge}: for each test profile, every same-disease training profile at Hamming distance $\leq1$ is removed. A test profile is retained only if at least one same-disease training profile remains after purging. Across seeds, the challenge contains 113 test profiles and 513--516 training profiles, with 96--99 training profiles removed. The distance constraint is applied only to same-disease train--test pairs.

All robustness experiments use the representative outer split with seed 42 and 50 independent corruption draws per level. For missing-symptom corruption, each observed symptom is retained independently with probability $\rho\in\{1.0,.9,.8,.7,.6,.5,.4,.3\}$. For spurious-symptom corruption, each absent symptom is independently activated with probability $a\in\{0,.0025,.005,.01,.02\}$. The plotted bands show variability across the 50 corruption draws.

\subsection{Reproducibility Protocol}
For each primary run, outer-split IDs, per-record predictions, selected hyperparameters with inner-cross-validation scores, and full grid-search results are logged. Run manifests record seed and model coverage, while purged-split membership files, mechanism and sensitivity results, robustness draws, and faithfulness scores are stored as secondary artifacts. Aggregation verifies that all ten required seeds are present before any primary table or figure is generated. Ten seeds are used for the primary, mechanism, and purged analyses; five seeds are used for balancing controls; seed 42 is used for robustness and explanation analyses.

\section{Results and Analysis}
\label{sec:results}

\subsection{Primary Classification Results}
The controlled $2\times2$ experiment is reported in Table~\ref{tab:main}. For BNB, accuracy is 60.78\% with raw untuned training and rises to 96.12\% after tuning alone. Adding Bernoulli augmentation with matched tuning increases accuracy by a further 1.38 percentage points to 97.50\%. This BNB gain is not significant after Holm correction ($p=0.1406$); the corresponding macro-F1 comparison is also not significant ($p=0.0703$). In a seed 42 diagnostic, BNB raw accuracy increases from 59.48\% to 96.55\% when $\alpha$ is reduced from 1 to 0.01, demonstrating the strong effect of smoothing in this sparse 85-class setting.

The largest augmentation gains occur for the boosted-tree models. XGBoost improves from 77.41\% to 95.52\% (+18.10 points), LightGBM from 66.81\% to 95.26\% (+28.45), and CatBoost from 85.78\% to 95.86\% (+10.09). All three accuracy gains remain significant after Holm correction ($p=0.0234$); their macro-F1 gains are also significant after correction. LightGBM is substantially more stable, with the accuracy SD decreasing from 20.94 to 1.24 points. Under Bernoulli-tuned training, each voting ensemble achieves 97.16\% accuracy, below BNB, so ensemble fusion does not exceed the best single learner on this benchmark.

\begin{table*}[!t]
\centering
\caption{Controlled $2\times2$ experiment over ten repeated real-profile holdouts. Accuracy and macro-F1 are percentages; tuned accuracy columns also report SD. U/T denote untuned/tuned, and Bern denotes Bernoulli augmentation. ``F1 R/B'' is raw-tuned/Bernoulli-tuned macro-F1. $\Delta$ and Holm-$p$ compare Bernoulli-tuned with raw-tuned accuracy. Deltas are computed before rounding, so subtraction of displayed means can differ by 0.01 point.}
\label{tab:main}
\scriptsize
\setlength{\tabcolsep}{3.5pt}
\begin{tabular}{@{}lccccccc@{}}
\toprule
\textbf{Model} & \textbf{Raw U} & \textbf{Raw T} & \textbf{Bern U} & \textbf{Bern T} & \textbf{F1 R/B} & \textbf{$\Delta$ pp} & \textbf{Holm-$p$}\\
\midrule
BNB & 60.78 & 96.12$\pm$1.36 & 97.24 & \textbf{97.50$\pm$0.49} & 93.4/95.7 & +1.38 & .1406\\
LR & 93.10 & 96.72$\pm$1.27 & 97.24 & 97.41$\pm$0.91 & 94.5/95.5 & +0.69 & .5625\\
MLP & 96.29 & 95.95$\pm$1.00 & 97.07 & 97.16$\pm$1.08 & 93.2/95.3 & +1.21 & .3750\\
Ens-Hard & 94.40 & 96.38$\pm$1.27 & 97.33 & 97.16$\pm$1.00 & 93.8/95.2 & +0.78 & .5469\\
Ens-Weighted & 96.47 & 96.64$\pm$1.25 & 97.16 & 97.16$\pm$0.91 & 94.4/95.2 & +0.52 & 1.0000\\
Ens-Soft & 96.03 & 96.47$\pm$1.31 & 97.16 & 97.16$\pm$0.91 & 94.1/95.2 & +0.69 & .7031\\
Nearest Proto. & 96.64 & 96.64$\pm$0.86 & 96.72 & 96.72$\pm$0.79 & 94.4/94.4 & +0.09 & 1.0000\\
KNN & 91.38 & 96.72$\pm$1.90 & 96.47 & 96.47$\pm$1.11 & 94.7/94.3 & -0.26 & 1.0000\\
RF & 94.74 & 95.00$\pm$1.62 & 95.69 & 95.60$\pm$1.25 & 91.9/92.9 & +0.60 & 1.0000\\
CatBoost & 85.78 & 85.78$\pm$1.83 & 96.12 & 95.86$\pm$1.62 & 77.6/93.2 & \textbf{+10.09} & \textbf{.0234}\\
XGBoost & 77.41 & 77.41$\pm$1.76 & 95.60 & 95.52$\pm$1.45 & 67.6/93.2 & \textbf{+18.10} & \textbf{.0234}\\
LightGBM & 67.24 & 66.81$\pm$20.94 & 95.09 & 95.26$\pm$1.24 & 56.7/93.1 & \textbf{+28.45} & \textbf{.0234}\\
\bottomrule
\end{tabular}
\end{table*}

\subsection{Template Proximity and Purged Generalization}
Train--test template proximity is high across the standard repeated holdouts. Among the 1,160 held-out evaluation instances, 14 are identical to a training profile, 1,042 differ by one bit, 98 differ by two bits, and only 6 differ by three or more bits. Thus, 91.0\% of held-out instances are within one bit and 99.5\% are within two bits of the training data. A simple Hamming 1-NN lookup already achieves $95.17\pm1.48$\% accuracy. Within diseases, the mean normalized Hamming distance is 0.0138, compared with 0.0595 between diseases, indicating tight local clusters of same-disease profiles.

\begin{figure}[!htbp]
\centering
\includegraphics[width=.98\columnwidth]{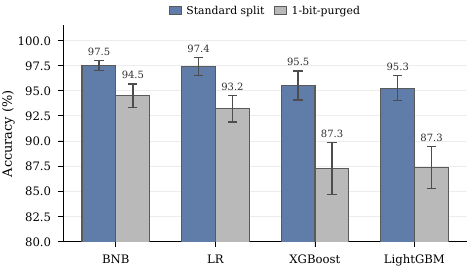}
\caption{Bernoulli-tuned accuracy on the standard split and the same-disease 1-bit-purged challenge. Bars show mean accuracy over ten seeds, error bars denote $\pm$1 SD, and the labels above the bars report the corresponding mean values.}
\label{fig:purgedplot}
\end{figure}

Figure~\ref{fig:purgedplot} compares Bernoulli-tuned accuracy on the standard split and the 1-bit-purged challenge for BNB, LR, XGBoost, and LightGBM. On the purged challenge, Hamming 1-NN drops to $86.81\pm1.47$\%, whereas Bernoulli-tuned BNB reaches 94.51\%, LR 93.19\%, XGBoost 87.26\%, and LightGBM 87.35\%. Augmentation still produces large gains for XGBoost (+16.02 points) and LightGBM (+30.53 points), both with Holm-corrected $p=0.0078$, whereas the BNB and LR gains remain below one percentage point. Purging also removes roughly 98 training profiles per seed, so the challenge is a deliberately stricter stress test rather than a direct estimate of performance after removing only exact duplicates.

\begin{table}[!htbp]
\centering
\caption{Same-disease 1-bit-purged challenge over ten seeds. Accuracy is given as mean$\pm$SD (\%). $\Delta$ compares Bernoulli-tuned and raw-tuned training on the same purged split; Holm-$p$ refers to accuracy.}
\label{tab:purged}
\scriptsize
\setlength{\tabcolsep}{3.0pt}
\begin{tabular}{@{}lcccc@{}}
\toprule
\textbf{Model} & \textbf{Raw} & \textbf{Bernoulli} & \textbf{$\Delta$ pp} & \textbf{Holm-$p$}\\
\midrule
BNB & 93.63$\pm$2.12 & 94.51$\pm$1.17 & +0.88 & .2500\\
LR & 92.48$\pm$1.57 & 93.19$\pm$1.32 & +0.71 & .2500\\
XGBoost & 71.24$\pm$1.88 & 87.26$\pm$2.58 & \textbf{+16.02} & \textbf{.0078}\\
LightGBM & 56.81$\pm$3.00 & 87.35$\pm$2.09 & \textbf{+30.53} & \textbf{.0078}\\
\bottomrule
\end{tabular}
\end{table}

\begin{figure}[!htbp]
\centering
\includegraphics[width=.98\columnwidth]{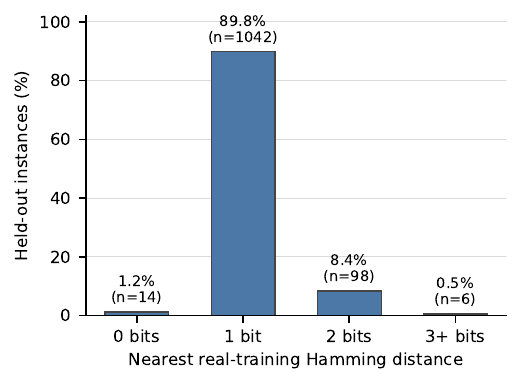}
\caption{Hamming distance to the nearest training profile for the 1,160 standard held-out evaluation instances.}
\label{fig:proximity}
\end{figure}

\subsection{Augmentation Mechanism and Rebalancing Controls}
For the mechanism ablation, hyperparameters are fixed for each model and only the training-data transformation is changed. As Table~\ref{tab:mechanism} indicates, removal-only perturbation performs similarly to full Bernoulli for BNB, LR, XGBoost, and CatBoost, whereas addition-only perturbation is less effective. Mild random removal of observed symptoms therefore appears to account for much of the regularization benefit. For example, XGBoost rises from 77.41\% on raw training to 95.43\% with removal-only perturbation and 95.78\% with the full Bernoulli rule.

LightGBM is the main exception. When the raw-tuned hyperparameters are held fixed and full Bernoulli augmentation is applied in the mechanism ablation, accuracy is only 24.57\%; with condition-specific tuning in Table~\ref{tab:main}, Bernoulli-tuned accuracy reaches 95.26\%. This large gap indicates strong sensitivity to the interaction between augmentation and LightGBM hyperparameters, reinforcing the need for matched tuning across training conditions.

\begin{table}[!htbp]
\centering
\caption{Mechanism ablation: mean accuracy (\%) over ten seeds, with raw-tuned hyperparameters held fixed for all variants.}
\label{tab:mechanism}
\scriptsize
\setlength{\tabcolsep}{3.0pt}
\begin{tabular}{@{}lccccc@{}}
\toprule
\textbf{Model} & \textbf{Raw} & \textbf{Exact dup.} & \textbf{Removal} & \textbf{Addition} & \textbf{Full}\\
\midrule
BNB & 96.12 & 95.26 & \textbf{97.67} & 93.71 & \textbf{97.67}\\
LR & 96.72 & 95.00 & \textbf{97.41} & 95.43 & 97.33\\
XGBoost & 77.41 & 68.28 & 95.43 & 71.81 & \textbf{95.78}\\
LightGBM & 66.81 & 68.36 & \textbf{71.72} & 51.72 & 24.57\\
CatBoost & 85.78 & 90.17 & \textbf{96.21} & 91.12 & 96.12\\
\bottomrule
\end{tabular}
\end{table}

A separate five-seed balancing analysis tests whether the gains can be explained by class frequency alone; each condition is optimized independently using the same leakage-controlled inner cross-validation protocol. BNB achieves mean accuracy of 96.55\% on raw data, 96.21\% with exact duplication, 96.72\% with ROS, 93.97\% with SMOTEN, and 97.59\% with Bernoulli training. The corresponding LR values are 96.38\%, 96.72\%, 97.24\%, 94.14\%, and 97.41\%. The separation is clearest for LightGBM: raw 74.31\%, exact duplication 76.72\%, ROS 74.31\%, class weighting 71.72\%, SMOTEN 74.83\%, and Bernoulli 95.52\%. Thus, the gain cannot be attributed to sample count or class balance alone. The exact-duplication results differ from Table~\ref{tab:mechanism} because this balancing analysis retunes each condition independently, whereas the mechanism ablation holds the raw-tuned hyperparameters fixed.

\subsection{Robustness and Explanation Faithfulness}
\begin{figure}[!htbp]
\centering
\includegraphics[width=.98\columnwidth]{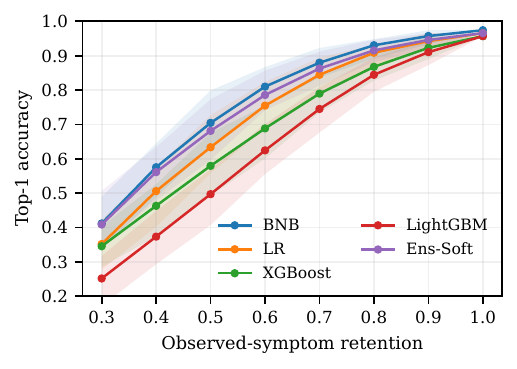}
\caption{Robustness of Bernoulli-tuned models to missing symptoms on seed 42. Curves give mean accuracy over 50 independent corruption draws, and bands give 95\% empirical intervals across the draws.}
\label{fig:robust-missing}
\end{figure}

\begin{figure}[!htbp]
\centering
\includegraphics[width=.98\columnwidth]{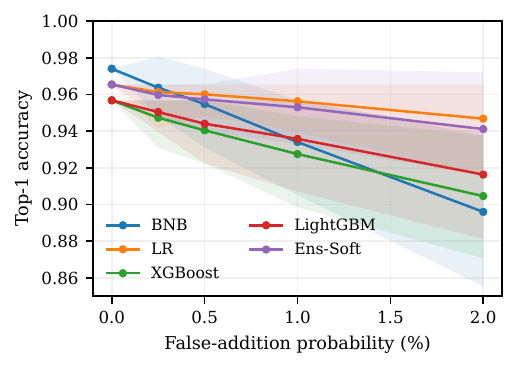}
\caption{Robustness of Bernoulli-tuned models to spurious symptoms on seed 42, using the same 50-draw protocol. Bands give 95\% empirical intervals across corruption draws.}
\label{fig:robust-spurious}
\end{figure}

Figures~\ref{fig:robust-missing} and~\ref{fig:robust-spurious} show the robustness of the Bernoulli-tuned models on the representative outer split with seed 42. When only 30\% of the originally observed symptoms are retained, accuracy is 41.2\% for BNB, 40.9\% for Ens-Soft, 35.2\% for LR, 34.6\% for XGBoost, and 25.2\% for LightGBM. At a 2\% false-positive symptom rate, LR reaches 94.7\%, Ens-Soft 94.1\%, LightGBM 91.6\%, XGBoost 90.5\%, and BNB 89.6\%. Because the model ranking changes between symptom deletion and false addition, robustness depends on both the model family and the type of input corruption.

BNB gives an exact per-feature decomposition. For predicted class $a$ and runner-up class $b$, feature $j$ contributes
\begin{equation}
\Delta_j=x_j\log\frac{\theta_{aj}}{\theta_{bj}}+(1-x_j)\log\frac{1-\theta_{aj}}{1-\theta_{bj}},
\label{eq:evidence}
\end{equation}
where $\theta_{cj}=P(x_j=1\mid y=c)$. To test faithfulness, we delete the top-$k$ active symptoms and compare the resulting drop in predicted-class probability with that caused by random deletion. In a complementary insertion test, we activate the top-$k$ symptoms in an all-zero profile and compare the resulting probability with random insertion.

\begin{table}[!htbp]
\centering
\caption{Faithfulness of BNB explanations on held-out profiles (seed 42). All values are means; Del. and Ins. denote deletion and insertion, and rand. denotes random selection.}
\label{tab:xai}
\scriptsize
\setlength{\tabcolsep}{3.5pt}
\begin{tabular}{@{}ccccc@{}}
\toprule
\textbf{$k$} & \textbf{Del. top} & \textbf{Del. rand.} & \textbf{Ins. top} & \textbf{Ins. rand.}\\
\midrule
1 & .102 & .065 & .268 & .193\\
2 & .335 & .201 & .701 & .634\\
3 & .592 & .431 & .882 & .862\\
5 & .908 & .784 & .989 & .961\\
\bottomrule
\end{tabular}
\end{table}

For every tested $k$, the probability of the predicted class decreases when the top-$k$ symptoms are removed and increases when the top-$k$ symptoms are inserted. These results support the faithfulness of the explanations to the fitted BNB model, but they do not establish the clinical validity of the learned disease--symptom relationships.

\subsection{Parameter Sensitivity and Statistical Reliability}
The selected augmentation setting is stable within the tested parameter range. Table~\ref{tab:sensitivity} reports mean inner-cross-validation macro-F1 over seeds 11, 22, and 33 for each parameter value. For both BNB and LR, the highest mean score occurs at $p=.90$, $q=.005$, and $n=50$. Scores vary only slightly across the tested $p$ and $q$ values. For BNB, increasing $n$ from 10 to 50 raises macro-F1 from 0.947 to 0.976, while $n=100$ gives 0.974; LR shows the same optimum at $n=50$ (0.972) before decreasing to 0.968 at $n=100$. Overall, the tested range suggests that moderate local perturbation with $n=50$ is preferable to either sparse augmentation or the largest tested synthetic expansion.

\begin{table}[!htbp]
\centering
\caption{Sensitivity analysis varying one factor at a time. Entries are mean inner-cross-validation macro-F1 over three seeds.}
\label{tab:sensitivity}
\scriptsize
\setlength{\tabcolsep}{4.0pt}
\begin{tabular}{@{}lccc@{}}
\toprule
\textbf{Axis} & \textbf{Value} & \textbf{BNB} & \textbf{LR}\\
\midrule
$p$ & .80 & .971 & .968\\
    & .85 & .974 & .970\\
    & \textbf{.90} & \textbf{.976} & \textbf{.972}\\
    & .95 & .972 & .957\\
\midrule
$q$ & 0 & .974 & .969\\
    & .0025 & .975 & .971\\
    & \textbf{.005} & \textbf{.976} & \textbf{.972}\\
    & .01 & .973 & .969\\
\midrule
$n$ & 10 & .947 & .957\\
    & 25 & .966 & .966\\
    & \textbf{50} & \textbf{.976} & \textbf{.972}\\
    & 100 & .974 & .968\\
\bottomrule
\end{tabular}
\end{table}

McNemar tests are carried out on all ten splits and reported descriptively; Bernoulli training is preferred at $p<.05$ on all ten splits for XGBoost and LightGBM and on 9 of 10 splits for CatBoost, but on none of the splits for BNB or LR. The primary inferential result remains the Holm-adjusted paired Wilcoxon analysis. On the representative seed 42 split, Bernoulli-tuned BNB correctly classifies 113 of 116 profiles (97.41\%), with a Wilson 95\% interval of 92.67--99.12\%. The large gains of the boosted-tree models further show that model rankings can depend strongly on the training condition.

\subsection{Limitations and Interpretation}
The variability of the repeated holdouts is not indicative of uncertainty across independent cohorts, as the repeated holdouts are overlapping. This purged challenge has a tighter same-disease separation constraint, and also reduces the real training data. The source records are curated binary profiles instead of patient episodes, and lack severity, duration, timing, demographics, comorbidity or negation information. Clinician adjudication, prospective evaluation, calibration-for-action analysis and external patient-level validation are also lacking. As with any other clinical prediction model, standard-split accuracy is not a measure of diagnostic safety or readiness for deployment, but rather a benchmark performance.

\FloatBarrier
\section{Conclusion and Future Work}
\label{sec:conclusion}
Under leakage-controlled evaluation, the benefit of asymmetric Bernoulli symptom perturbation is strongly model dependent. Mean BNB accuracy on the standard splits is 97.50\%, representing a small gain over the tuned raw baseline that is not statistically significant after Holm correction. XGBoost, LightGBM, and CatBoost, by contrast, show large and statistically significant gains. Conventional rebalancing does not reproduce the LightGBM improvement, and the mechanism ablation suggests that mild symptom removal accounts for much of the regularization effect in several models.

These headline results should be interpreted together with template proximity. Most standard held-out profiles lie within one symptom bit of the training data, and a simple Hamming 1-NN baseline already achieves 95.17\% accuracy. Under the more demanding 1-bit-purged challenge, overall accuracy decreases, yet XGBoost and LightGBM retain significant augmentation benefits. Future work should evaluate independent patient-level Bangla datasets, incorporate temporal, demographic, and other contextual features, investigate correlation-aware discrete generators, assess calibration, and perform external cross-dataset validation in line with contemporary reporting guidance~\cite{tripod_ai}. Split IDs, predictions, hyperparameter searches, and diagnostic artifacts are released to support reproduction and independent assessment.

\FloatBarrier
\renewcommand{\IEEEbibitemsep}{2.4pt plus 0.3pt minus 0.15pt}

\end{document}